\documentclass[11pt, a4paper]{deepmind}
\pdftrailerid{redacted}

\usepackage[utf8]{inputenc}
\usepackage{kantlipsum, lipsum}
\usepackage{dsfont}
\usepackage{dm-colors}

\usepackage[authoryear, sort&compress, round]{natbib} 
\usepackage{caption}
\usepackage{subcaption}
\usepackage{wrapfig}
\usepackage{hyperref}
\usepackage{tikz}
\usepackage{pifont}
\usepackage{float}
\usepackage{svg}

\newcommand{\xmark}{\ding{55}}
\newcolumntype{P}[1]{>{\centering\arraybackslash}p{#1}}
\graphicspath{{figures/}}

\title{AndroidEnv: A Reinforcement Learning Platform for Android}

\correspondingauthor{\texttt{[kenjitoyama,hamelphi,agergely,gcomanici]@deepmind.com}}

\author[*,1]{Daniel Toyama}
\author[*,1]{Philippe Hamel}
\author[*,1]{Anita Gergely}
\author[*,1]{Gheorghe Comanici}
\author[1]{Amelia Glaese}
\author[1]{Zafarali Ahmed}
\author[1]{Tyler Jackson}
\author[1]{Shibl Mourad}
\author[1]{Doina Precup}

\affil[*]{Equal contributions}
\affil[1]{DeepMind}

\begin{abstract}
We introduce AndroidEnv, an open-source platform for Reinforcement Learning (RL) research built on top of the Android ecosystem. AndroidEnv allows RL agents to interact with a wide variety of apps and services commonly used by humans through a universal touchscreen interface. Since agents train on a realistic simulation of an Android device, they have the potential to be deployed on real devices. In this report, we give an overview of the environment, highlighting the significant features it provides for research, and we present an empirical evaluation of some popular reinforcement learning agents on a set of tasks built on this platform.
\end{abstract}

\begin{document}

\maketitle

\newcommand{\expect}[2]{\mathds{E}_{{#1}} \left[ {#2} \right]}
\newcommand{\myvec}[1]{\boldsymbol{#1}}
\newcommand{\myvecsym}[1]{\boldsymbol{#1}}
\newcommand{\vx}{\myvec{x}}
\newcommand{\vy}{\myvec{y}}
\newcommand{\vz}{\myvec{z}}
\newcommand{\vtheta}{\myvecsym{\theta}}

\section{Introduction}
Reinforcement learning (RL) is a branch of artificial intelligence (AI) which studies computational models of learning from interaction with an environment and from numerical rewards~\citep{sutton2018}. RL methods have demonstrated success not only in game playing, for example checkers~\citep{Schaeffer92aworld}, chess~\citep{CAMPBELL200257, silver2017-alpha_zero}, Go~\citep{silver2016mastering}, poker~\citep{MoravcikSBLMBDW17}, Atari~\citep{mnih2015humanlevel} and Starcraft II~\citep{journals/nature/VinyalsBCMDCCPE19}, but also in real-world applications, such as robotics~\citep{Kormushev2013ReinforcementLI}, logistics~\citep{REFANDIDIS2001}, chemical synthesis~\citep{ journals/nature/SeglerPW18} and  personalised recommendations~\citep{liu2019deep}. In many of these applications, RL agents were able to achieve super-human performance, yet they can be prone to over-specialising to any single domain.
In order to assess the performance of RL algorithms over a range of different tasks, it is desirable to have platforms which expose diverse challenges through a unified interface. This approach was pioneered in the original Atari suite~\citep{Bellemare_2013} and has been followed up by a variety of platforms, such as DeepMind Lab~\citep{beattie2016dmlab}, OpenAI Universe~\citep{open_ai_universe} and World of Bits~\citep{liu2018reinforcement}. To complement these existing platforms, we present \emph{AndroidEnv}, a research platform built on top of the Android Operating System (OS). The open-source library, along with detailed technical documentation and a set of tasks are available on \href{https://github.com/deepmind/android_env}{GitHub}.\footnote{
https://github.com/deepmind/android\textunderscore env}

AndroidEnv has a universal touchscreen interface that enables the empirical evaluation of general purpose RL algorithms designed to tackle a wide variety of tasks. The agent-environment interaction in AndroidEnv matches that of a user and a real device: the screen pixels constitute the observations, the action space is defined by touchscreen gestures, the interaction is real-time, and actions are executed asynchronously, while the environment runs at its own time scale. With these features, agent performance can be realistically compared to humans. Moreover, environments that behave as closely as possible to their real-world counterparts also facilitate production deployment, without added work to adapt to different interfaces or data distributions.

We chose Android as the underlying system because it is a popular, open-source operating system with over two billion monthly active users and a selection of over two million applications. The sheer number of applications, built for a multitude of important aspects of human life, ranging from education and business to communication and entertainment, provides virtually unlimited challenges for RL research. Furthermore, externally written apps ground the research in real problems, avoiding common pitfalls of systems tailored for specific research agendas.

This technical report is structured as follows: Section~\ref{sec:rl_env} provides an overview of the notable features of AndroidEnv. Section~\ref{sec:tasks} describes what defines a \emph{Task}, and presents a set of tasks included in the release. Section~\ref{sec:baselines} provides some initial empirical results of popular RL agents on a selection of AndroidEnv tasks. Section~\ref{sec:technical} provides some technical details worth considering when using the AndroidEnv platform. Lastly, Section~\ref{sec:rl} discusses some existing RL research platforms and highlights their relevance to AndroidEnv.

\section{Environment Features} \label{sec:rl_env}

AndroidEnv enables RL agents to interact with, and learn to solve tasks on any Android application, including the operating system itself. In particular, AndroidEnv implements the dm\_env API~\citep{dm_env2019} on top of an emulated Android device. Virtual, emulated Android devices allow the dynamics of the environment to be entirely generated by the OS itself. In the rest of this section, we expand on the most important distinguishing features of the environment.

\subsection{Real-time execution}

The Android OS, whether it is running as an emulator or on a real device, runs in real-time, independently of the agent interacting with it. All observations and actions are asynchronous and OS does not pause when providing observations or when accepting actions. Users can control the rates for fetching observations and for sending actions, but they cannot speed up or slow down the OS. As such, AndroidEnv is unable to run in lock-step, and agents may need to handle a non-negligible amount of delay between consecutive action executions. Furthermore, the screen refresh rate varies between 60Hz and 120Hz and capturing the screen beyond that limit does not provide the agent with more information. Android and its specific apps are in control of processing and interpreting agent actions, and the platform allows buffering up to a device and version-dependent limit. However, sending a high number of actions at a time does not give the agent more control over the simulation. These characteristics make AndroidEnv a more naturalistic platforms for developing and testing RL algorithms.

\subsection{Action interface}

\begin{wrapfigure}{r}{0.45\textwidth}
    \vspace{-20pt}
	\begin{center}
	    \includegraphics[width=0.45\textwidth]{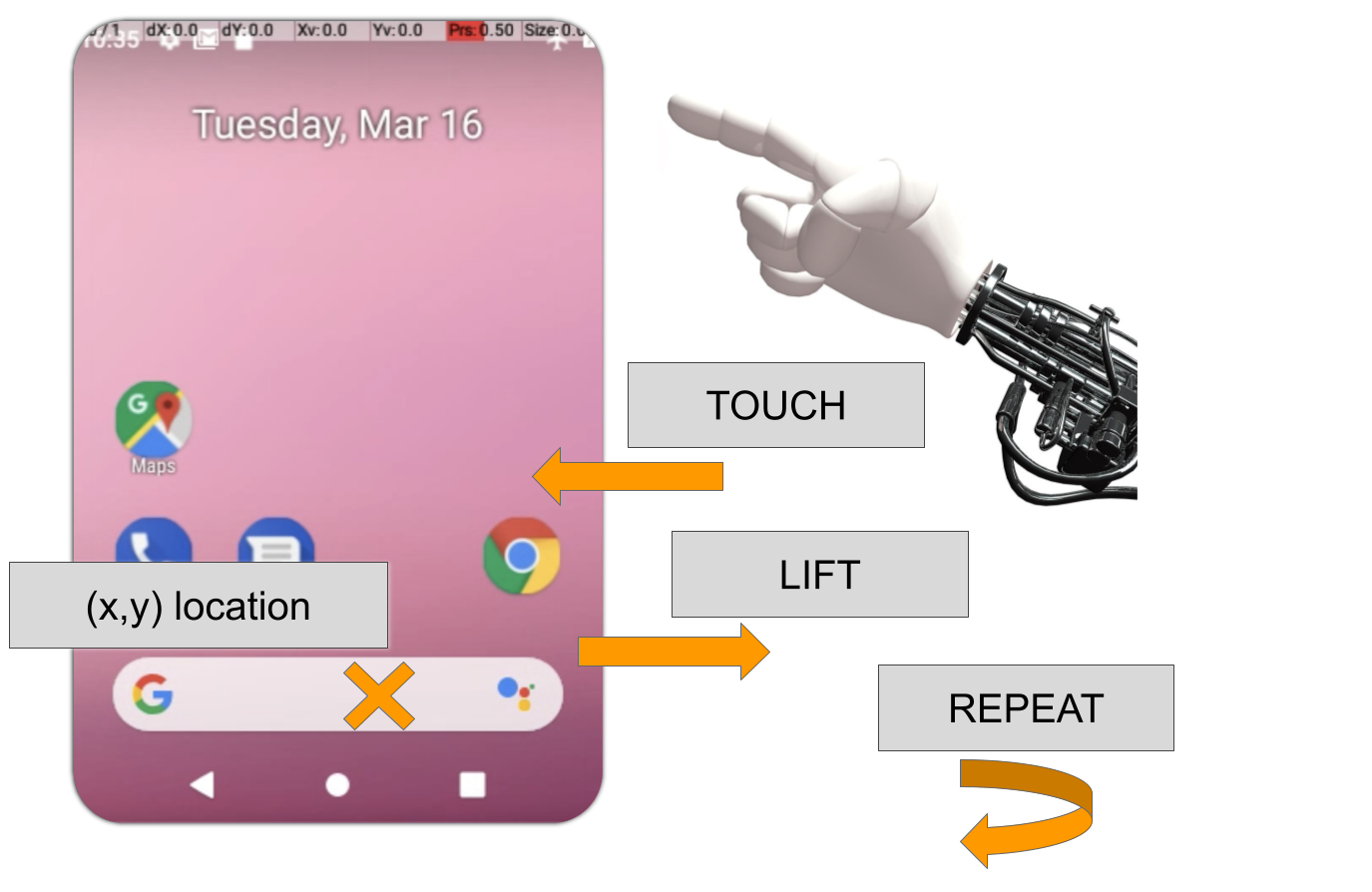}
	\end{center}
	\vspace{-10pt}
	\caption{The action space is composed of a discrete action type and a screen location.}
	\label{fig:action_space}
\end{wrapfigure}

\paragraph{Raw action space.} The native action space of the environment consists of a tuple consisting of a position $ (x,y) \in [0,1]\times[0,1]$, determining the location of the action on the screen, and a discrete value $\verb+ActionType+ \in \{\verb+TOUCH+,\ \verb+LIFT+,\ \verb+REPEAT+\}$ indicating whether the agent opts for touching the screen at the indicated location, lifting the pointer from the screen, or repeating the last chosen action, respectively. This action space is the same across all tasks and apps. 

It is worth noting that while two actions $a_1 = \{\verb+ActionType+=\verb LIFT, \verb+ position+=(x_1, y_1)\}$ and $a_2 = \{\verb+ActionType+=\verb LIFT, \verb+ position+=(x_2, y_2)\}$ are different from the agent's perspective, in practice they result in the same effect on the device, because the \emph{lack} of a touch has no association to  a particular location.

\paragraph{Gestures.} The complexity of the interface arises from the fact that individual raw actions on their own do not necessarily trigger a meaningful change in the environment. It is more useful for agents to control Android applications via \emph{gestures}, such as pressing, long pressing, swiping, scrolling, or drag-and-drop. Each of these correspond to a particular sequence of raw actions: for example, a screen \emph{touch} at a particular location, followed by a \emph{lift} of the the imaginary finger is a sequence that Android can interpret as a \emph{press of a button}. Similarly, Android will interpret a sequence of aligned \emph{touches}  as \emph{scrolling}. 

\begin{figure}[h]
    \centering
    \begin{subfigure}{0.2\textwidth}
        \centering
        \includegraphics[width=\textwidth]{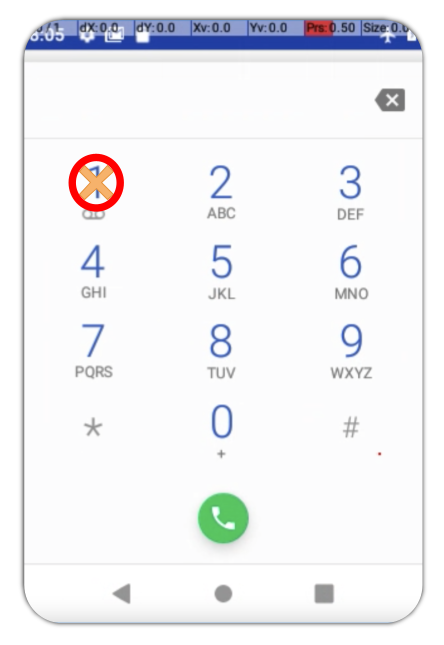}
        \caption{Tapping}
        \label{fig:tapping}
    \end{subfigure}
    \hfill
    \begin{subfigure}{0.2\textwidth}
        \centering
        \includegraphics[width=\textwidth]{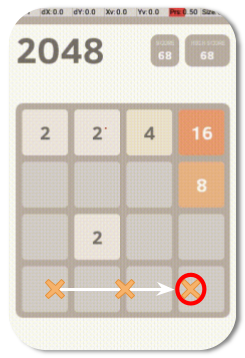}
        \caption{Swiping}
        \label{fig:swiping}
    \end{subfigure}
    \hfill
    \begin{subfigure}{0.2\textwidth}
        \centering
        \includegraphics[width=\textwidth]{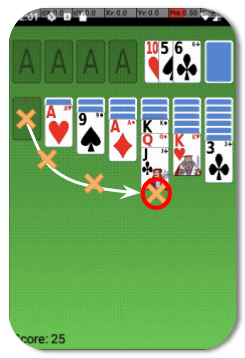}
        \caption{Drag-and-drop}
        \label{fig:drag_and_drop}
    \end{subfigure}
    \hfill
    \begin{subfigure}{0.25\textwidth}
        \centering
        \includegraphics[width=\textwidth]{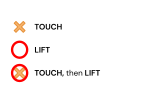}
        \label{fig:legend}
    \end{subfigure}
    \hfill
    \caption{Examples of gestures. Actions are performed one after the other, tracing out a particular path.}
    \label{fig:gestures}
\end{figure}

This distinction between the \emph{raw} action space and a particular app \emph{interface} makes AndroidEnv a challenging domain. A random sequence of actions will typically have a small probability of producing a meaningful gesture in most Android apps. This need to compose actions, paired with the difficulty of solving the underlying task itself, leads to a difficult exploration problem. For example, in order to learn to play chess in AndroidEnv, an agent must not only find a winning strategy, it also has to learn to move pieces through drag-and-drop gestures.

\paragraph{Relation to observations.} Another notable feature of AndroidEnv is the spatial correlation between actions and observations. Often, an action can result in local changes in the pixels near the location of the action, or the position of certain items in the observation might hint at the next best location to take an action. In particular, the screen is often suggestive of the kind of \emph{gestures} the application expects: smartphone users would often find it intuitive to \emph{tap} where they see an item in the shape of a button, or to \emph{scroll} where they see a drop-down menu.

\paragraph{Altering the action space.} AndroidEnv allows users to define wrappers around the raw action space of the environment. For example, one might discretise the action space by splitting up the screen into a grid, restrict the $\verb+ActionType+$ to $\verb+TOUCH+$, or group action sequences like $[\verb+LIFT+, \verb+TOUCH+, \verb+LIFT+]$ into a single \emph{tap} action.  We provide some useful and natural wrappers (see Section~\ref{sec:technical}). Note that these wrappers but alter the set of actions available to the agent,  but not the way in which AndroidEnv interprets raw actions.

\subsection{Observations}

\paragraph{Observation space.} The observation space of AndroidEnv consists of three main components: \{\verb+pixels+,\ \verb+timedelta+,\ \verb+orientation+\}. The most notable component is \verb+pixels+, representing the current frame as an RGB image array. Its dimensions will depend on the device used (real or virtual), but given that it will correspond to real device screen sizes, this array will typically be large (of course, users can scale down their dimensionality, e.g. with wrappers). The \verb+timedelta+ component captures the amount of time passed since AndroidEnv fetched the last observation. The \verb+orientation+, even though it does not affect the layout of the RGB image in the observation, might carry relevant information for the agent. For example, if there is text on the screen, its orientation is useful for automatic processing. As mentioned above, observations often carry spatial cues and are suggestive of meaningful gestures to perform in a given state. The fact that the observation space is the same across all tasks is what makes it useful for agents, and creates the opportunity to generalize across tasks.

\begin{wrapfigure}{r}{0.28\textwidth}
	\centering
	\vspace{-20pt}
	\includegraphics[width=0.28\columnwidth]{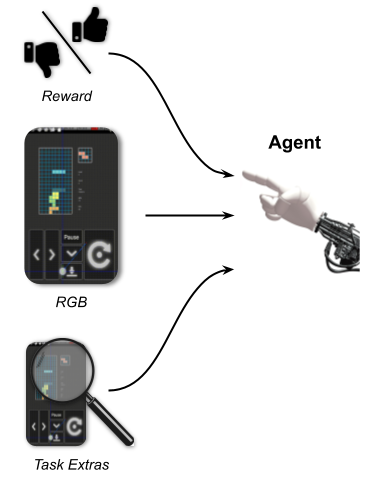}
	\vspace{-20pt}
	\caption{Information available to the agent.}
	\label{fig:observation}
\end{wrapfigure}

\paragraph{Task extras.} \label{Task extras}  In addition to default observations, (\{\verb+pixels+,\ \verb+timedelta+,\ \verb+orientation+\}), some  tasks might expose structured information after each step (see Sec.~\ref{sec:tasks}). An \emph{extra} in AndroidEnv is any information that the environment sends to aid the understanding of the task. The information sent through this channel is typically very useful for learning, yet difficult to extract from raw pixels. For example, extras may include signals indicating events such as a button press or opening of a menu, text displayed on the screen in string format, or a simple numerical representations of the displayed state. Note that extras are a standard mechanism for communicating information used in Android apps.

We note that, unlike the observation and raw action space, which are the same across all AndroidEnv, task extras are specific to individual tasks, are entirely optional, and may not be available at all. Furthermore, task extras, even if provided, are not part of the default observation; rather AndroidEnv returns them upon explicit request (see detailed documentation).

\section{Tasks} \label{sec:tasks}

While Android is an operating system with no inherent rewards or episodes, AndroidEnv provides a simple mechanism for defining \emph{tasks} on it. Tasks capture information such as episode termination conditions, rewards, or the apps with which the agent can interact. Together, these define a specific RL problem for the agent.

\begin{figure}[h]
    \centering
    \begin{subfigure}{0.18\textwidth}
        \centering
        \includegraphics[width=\textwidth]{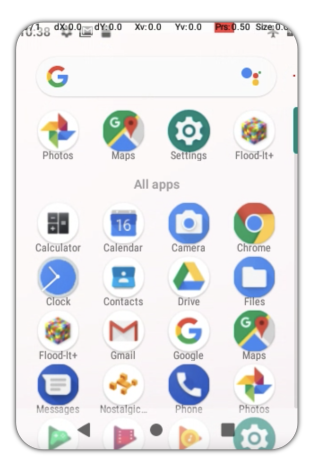}
        \caption{Android menu}
        \label{fig:menu}
    \end{subfigure}
    \hfill
    \begin{subfigure}{0.18\textwidth}
        \centering
        \includegraphics[width=\textwidth]{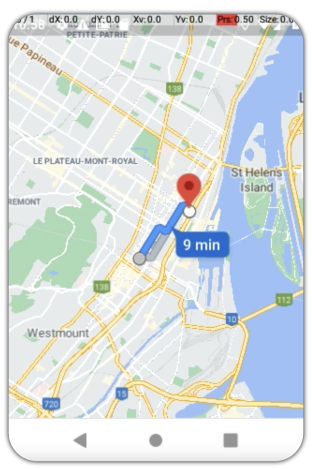}
        \caption{Google Maps}
        \label{fig:maps}
    \end{subfigure}
    \hfill
    \begin{subfigure}{0.18\textwidth}
        \centering
        \includegraphics[width=\textwidth]{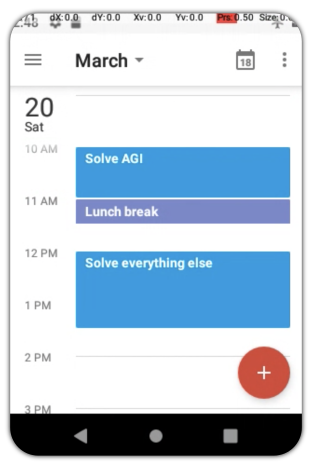}
        \caption{Calendar}
        \label{fig:calendar}
    \end{subfigure}
    \hfill
    \begin{subfigure}{0.18\textwidth}
        \centering
        \includegraphics[width=\textwidth]{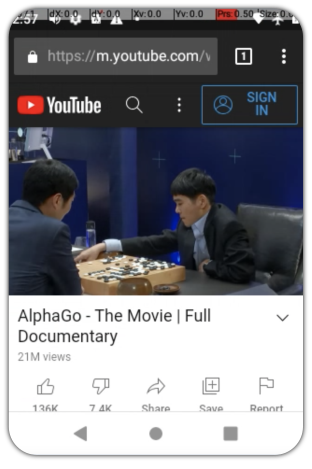}
        \caption{Chrome}
        \label{fig:chrome}
    \end{subfigure}
    \hfill
    \begin{subfigure}{0.18\textwidth}
        \centering
        \includegraphics[width=\textwidth]{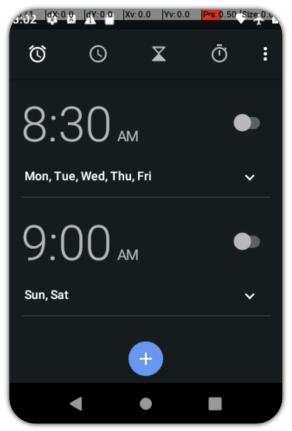}
        \caption{Clock}
        \label{fig:clock}
    \end{subfigure}
    \hfill
    \caption{Examples of Android OS apps and use cases.}
    \label{fig:beyondgames}
\end{figure}

\paragraph{Task structure.} We capture aspects that make up a task definition in a \emph{Task} protocol buffer message. These include information on:

\begin{itemize}
    \item How to initialise the environment: for example, installing particular applications on the device.
    \item When should an episode be reset: for example, upon receiving a particular message from the device or app, or upon reaching a certain time limit.
    \item Events triggered upon an episode reset: for example, launching a given app, clearing the cache, or pinning the screen to a single app (hence restricting the agent's interaction to that app).
    \item How to determine the reward: for example, this might depend on different signals coming from Android, such as Android accessibility service or log messages implemented in applications.
\end{itemize}

With these protocol buffer messages, users can define a wide variety of tasks on Android. For example, a task could be to set an alarm in the Android standard Clock app, by opening this app upon launch, and rewarding the agent and ending an episode once an alarm has been set. We detail the full specification of the protocol buffer message structure in the code repository.

\paragraph{Available tasks.} Along with the AndroidEnv platform implementation, we provide an initial set of ready-to-use tasks. At the time of the release, this includes over 100 tasks across roughly 30 different apps, ranging from basic tasks with straightforward objectives, to more sophisticated tasks that require long-term reasoning. The selection contains time-sensitive tasks (e.g. \texttt{catch}), physics-based environments (e.g. \texttt{vector\_pinball}), puzzles (e.g. \texttt{classic\_2048}), card games (e.g. \texttt{simple\_solitaire}), spatial reasoning (e.g. \texttt{perfection}), UI navigation (e.g. \texttt{clock\_set\_timer}), strategy games (e.g. \texttt{droidfish}) and more. Note that several of these tasks are defined around the same app by varying parameters such as the game level, the reward signal or the difficulty.
We emphasize that this set serves as a starting point and not as a definitive benchmark. Users can define their own tasks. We refer the reader to the code repository for instructions on creating additional tasks, as well as for an up-to-date list of available tasks.

\section{Experimental results} \label{sec:baselines}

In this section, we present some empirical results for a selection of baseline RL agents on a small subset of tasks. For our experiments, we used the Acme framework ~\citep{hoffman2020acme} and its TensorFlow~\citep{tensorflow2015-whitepaper} agents available at \href{https://github.com/deepmind/acme/tree/master/acme/agents/tf}{Acme's Github Repository}.\footnote{https://github.com/deepmind/acme/tree/master/acme/agents/tf}

Since the action interface in AndroidEnv is a hybrid of discrete and continuous components, we defined some \nameref{wrappers} (described below) for ease of experimentation. The continuous control agents we ran are Deep Deterministic Policy Gradient (DDPG)~\citep{LillicrapHPHETS15}, its distributional version (D4PG)~\citep{barthmarton2018d4pg}, and Maximum a Posteriori Policy Optimisation (MPO)~\citep{abdolmaleki2018mpo}. All these agents interact with a wrapped version of the environment for which they have to provide an $\verb+ActionType+$ as a continuous value in the interval $[0, 1]$. AndroidEnv rounds this number to the nearest integer and forwards the corresponding discrete \texttt{ActionType} to the simulator. 

We also tested the following agents designed for finite action interfaces: DQN~\citep{mnih2015humanlevel}, IMPALA~\citep{espeholt2018impala}, and R2D2~\citep{conf/iclr/KapturowskiOQMD19}. In this case, we discretised the screen as a $6 \times 9$ grid, resulting in $108$ possible actions, corresponding to a choice of $\verb+ActionType+$ among $(\verb+LIFT+, \verb+TOUCH+)$ combined with any of the $54$ cells in the grid. To help memoryless agents, we augmented the current observation with a one-hot encoding of the location of the last taken action, which provides a more informative input for learning.

For our experiments, we chose the following tasks: \texttt{catch}, \texttt{rocket sleigh}, \texttt{press button}, \texttt{apple flinger}, \texttt{2048}, \texttt{blockinger}. They were selected to be representative of the variety of apps, difficulties, and action interfaces available across Android. This variety is reflected in the experimental results, showing that the same agents can have drastically different performance depending on each of these factors. For example, most agents perform well on tasks such as \texttt{catch} that have a simple action interface and dense rewards, whereas the combination of a highly structured interface, time sensitivity and sparse rewards render \texttt{blockinger} particularly difficult to solve.

\begin{figure}[t]
    \centering
    \begin{subfigure}{0.15\textwidth}
        \centering
        \includegraphics[width=\textwidth]{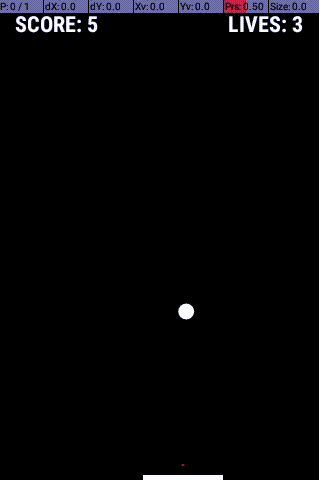}
        \caption{Catch}
        \label{fig:catch}
    \end{subfigure}
    \hfill
    \begin{subfigure}{0.15\textwidth}
        \centering
        \includegraphics[width=\textwidth]{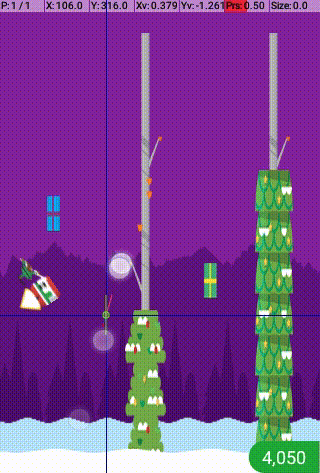}
        \caption{Rocket Sleigh}
        \label{fig:rocket_sleigh}
    \end{subfigure}
    \hfill
    \begin{subfigure}{0.15\textwidth}
        \centering
        \includegraphics[width=\textwidth]{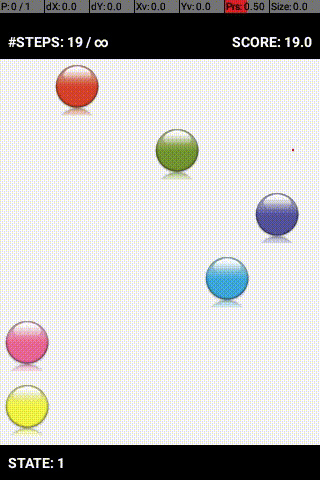}
        \caption{Press Button}
        \label{fig:mdp_flood_it}
    \end{subfigure}
    \hfill
    \begin{subfigure}{0.15\textwidth}
        \centering
        \includegraphics[width=\textwidth]{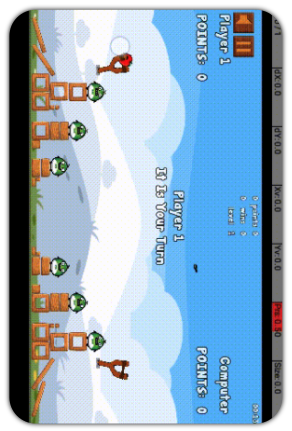}
        \caption{Apple Flinger}
        \label{fig:apple_flinger}
    \end{subfigure}
    \hfill
    \begin{subfigure}{0.15\textwidth}
        \centering
        \includegraphics[width=\textwidth]{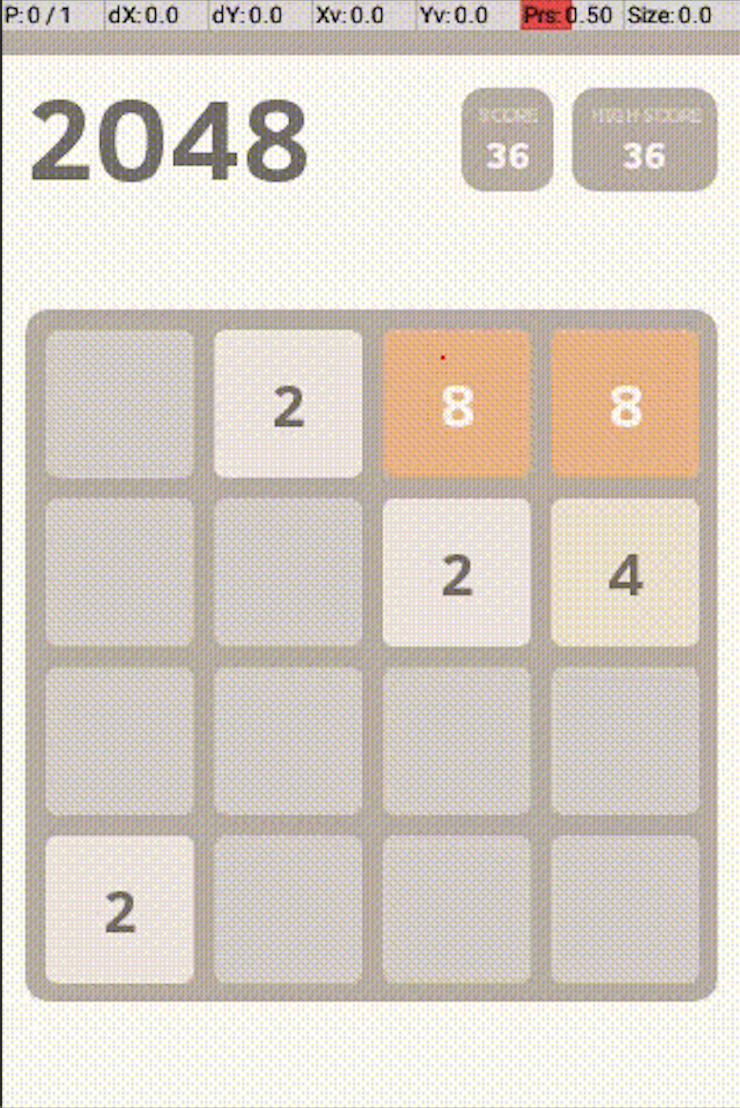}
        \caption{2048}
        \label{fig:2048}
    \end{subfigure}
    \hfill
    \begin{subfigure}{0.15\textwidth}
        \centering
        \includegraphics[width=\textwidth]{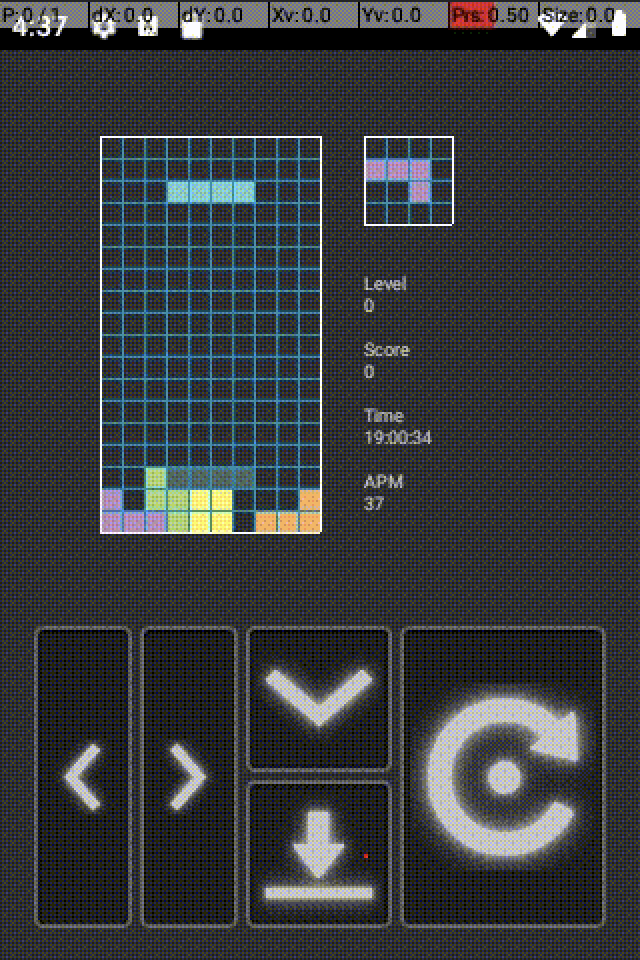}
        \caption{Blockinger}
        \label{fig:blockinger}
    \end{subfigure}
    \hfill
    \caption{Small selection of tasks used in the experiments.}
    \label{fig:beyondgames2}
\end{figure}

\begin{figure}[h]
\centering
\includegraphics[width=0.85\textwidth]{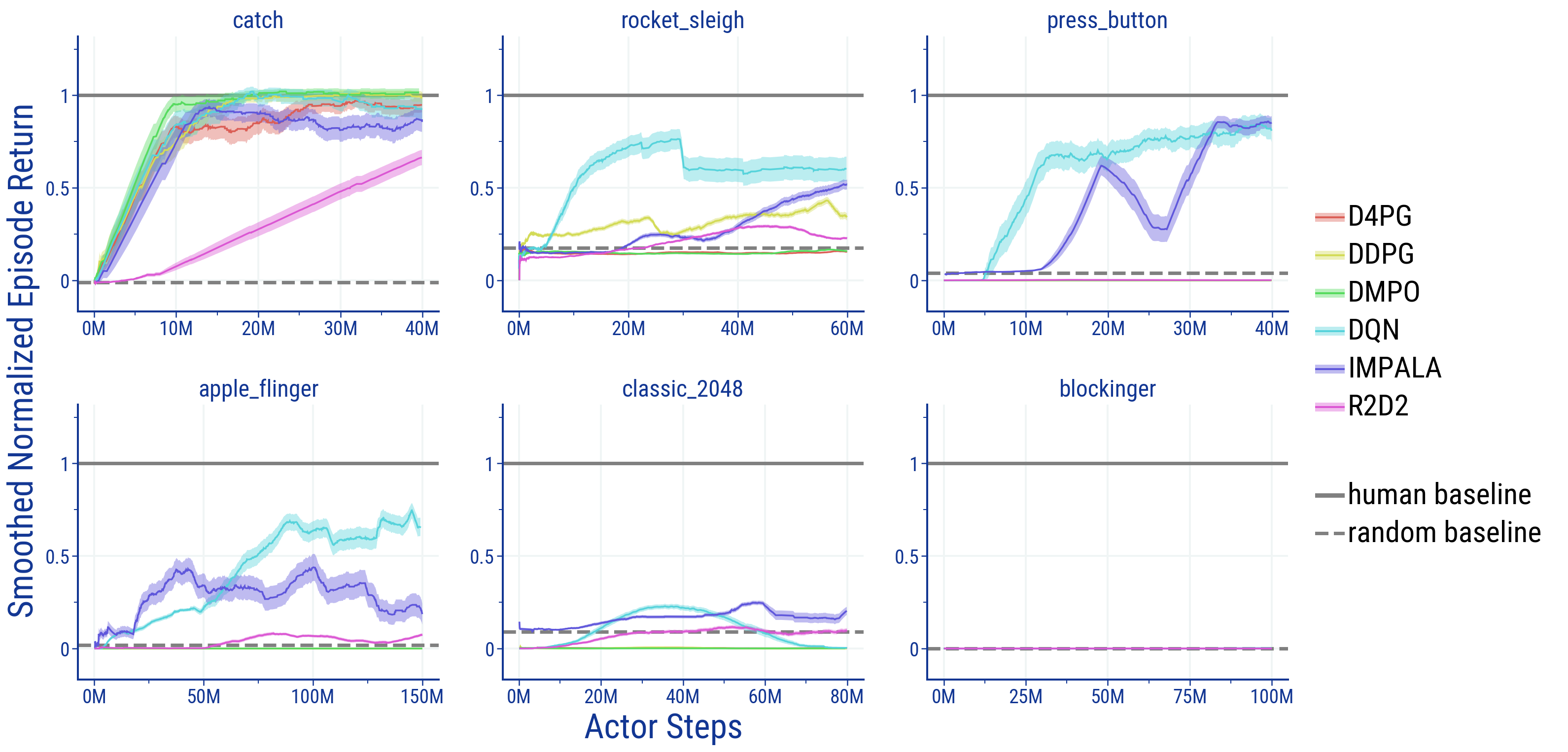}
\caption{\textbf{Agent performance}: The baseline continuous and discrete control agents ran on selection of AndroidEnv tasks, covering games where the action interface requires interactions including localised touches (\texttt{catch}), swiping (\texttt{classic\_2048}), and drag-and-drop (\texttt{apple\_flinger}). Continuous control agents perform well only in tasks where the interface does not expect complex gestures, but fail to achieve reasonable performance otherwise. Discrete control agents display better overall performance. We compiled the results above by averaging human-normalized scores (with \texttt{1.0} corresponding to average human performance) over four different seeds for each agent configuration. Note the clear difference in task difficulty, highlighted by the performance of baseline agents, with \texttt{catch} being solved by almost all agents, while no agents can generate useful behavior on \texttt{blockinger}.}  \label{fig:baselines}
\end{figure}

Since none of these tasks require high-resolution inputs to achieve optimal behavior, we down-sampled the image observation to $80 \times 120$ pixels. Since this size is comparable to the resolution commonly used in the ATARI Learning Environment, we were able to run all agents using the network architectures reported by the authors of each corresponding agent. We generated training data using 128 distributed actors and we compiled results for each hyper-parameter configuration by averaging the performance of 4 independent runs using different seeds. See Figure~\ref{fig:baselines} for an overview of the results of these experiments.

\section{Technical Details} \label{sec:technical}

\paragraph{ADB-based communication.} \emph{Android Debug Bridge} (ADB) provides a way of communicating with an Android device, be it physical or virtual. It exposes a shell that allows users to send commands to the device. AndroidEnv uses ADB for control operations, such as launching an app, querying the current activity, resetting episodes and listening for task extras coming from the app. 

\paragraph{Simulator.} AndroidEnv uses the Android Emulator\footnote{https://developer.android.com/studio/run/emulator}, which is provided with Android Studio as its default simulator. In order to run the emulator,  users need to specify an \emph{Android Virtual Device} (AVD). In particular, one can use  Android Studio to create AVDs with a specific screen resolution and OS version. Thus, users can choose the device type used for RL simulations. In principle, they can also extend AndroidEnv to work with other simulators. Simulations also provide a safe environment for RL agents to learn and make mistakes without any real world impact.

\paragraph{Real-time interaction.} Because AndroidEnv is a real-time platform, some timings will be inherently unpredictable. Depending on the machine and the simulator, there is a rate limit at which AndroidEnv fetches observations from the OS, which depends on the resolution of the device, the performance of the machine, and whether the rendering is done through software or hardware.

\begin{figure}[h]
    \centering
    \begin{tikzpicture}

\path [fill=blue!25] (0,-0.1) -- (0,1) -- (12,1) -- (12,-0.1) -- (1,-0.1);

\draw [thick, ->] (0,0) -- (12, 0) node[align=left, below] {time};

\begin{footnotesize}
\draw (1,-0.1) node[align=center, below]{$O_{t}$ \\ $A_t$} -- (1, 1) node[align=center, above] {Agent receives\\ observation and\\ sends action};

\draw (3.5,0) node[above, align=center]{Environment\\ steps forward};

\draw (6,-0.1) node[align=center, below]{$O_{t+1}$ \\ $A_{t+1}$} -- (6, 1) node[align=center, above] {Agent receives\\ observation and\\ sends action};

\draw (8.5,0) node[above, align=center]{Environment\\ steps forward};

\draw (11,-0.1) node[align=center, below]{$O_{t+2}$  \\ $A_{t+2}$} -- (11, 1) node[align=center, above] {Agent receives\\ observation and\\ sends action};
\end{footnotesize}

\draw [dotted] (1,-0.8) -- (1,-1) -- (6, -1) --(6, -0.8);
\draw (3.5,-1) node[below]{$\Delta t$};

\end{tikzpicture}
    \caption{Timeline of lockstep interaction between an environment and an agent. After sending an observation, the environment waits for the agent's action before stepping the simulation time forward.}
    \label{fig:timeline_lockstep}
\end{figure}
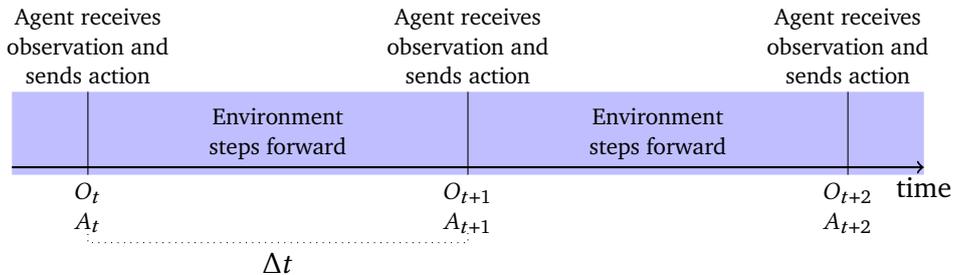

Another important factor to consider in real-time environments is that agents require some deliberation time to generate the next action, given an observation. In traditional lockstep environments, the environment generates an observation and pauses to wait until the agent responds with an action, before stepping the simulation forward, as illustrated in Figure~\ref{fig:timeline_lockstep}. Thus, in that setting, the actor deliberation time has no consequence on the agent-environment interaction. In a real-time setting, the environment does not pause to wait for the agent's action, as seen in Fig.~\ref{fig:timeline_lockstep}, so large deliberation times can be harmful to performance. We view this as an interesting challenge that RL agents need to tackle, and which is not present in most other simulation platforms.  

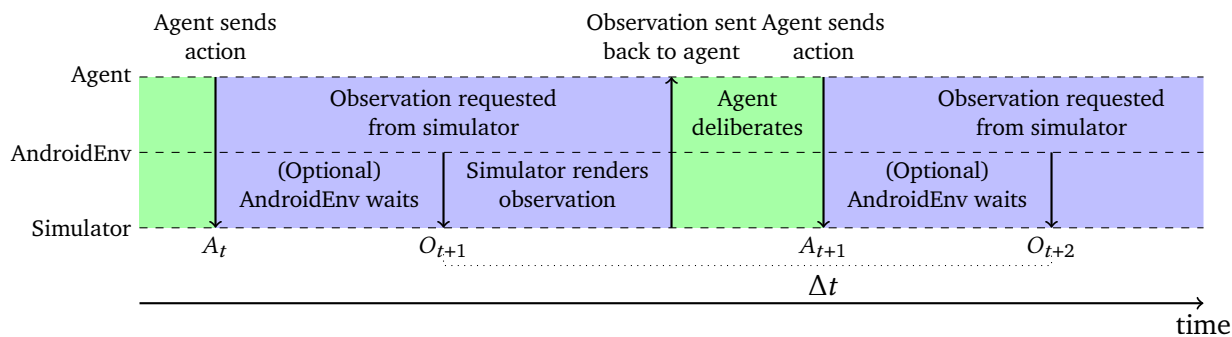
\begin{figure}[h]
    \centering
    \begin{tikzpicture}

\path [fill=green!35] (0,-1) -- (0,1) -- (1,1) -- (1,-1) -- (0,-1) ;
\path [fill=blue!25] (1,-1) -- (1,1) -- (7,1) -- (7,-1) -- (1,-1) ;
\path [fill=green!35] (7,-1) -- (7,1) -- (9,1) -- (9,-1) -- (7,-1) ;
\path [fill=blue!25] (9,-1) -- (9,1) -- (14,1) -- (14,-1) -- (10,-1) ;

\draw [thick, ->] (0,-2) -- (14, -2) node[below] {time};

\begin{footnotesize}

\draw [dashed] (0,1) node[left] {Agent} -- (14, 1);

\draw [dashed] (0,0) node[left] {AndroidEnv} -- (14, 0);

\draw [dashed] (0,-1) node[left] {Simulator} -- (14, -1);

\draw  [thick, <-] (1,-1) node[below]{$A_t$} -- (1, 1) node[align=center, above] {Agent sends \\ action\strut};

\draw (2.5,0) node[below, align=center]{(Optional)\\ AndroidEnv waits};

\draw  [thick, <-] (4,-1) node[below]{$O_{t+1}$} -- (4, 0) node[align=center, above] {Observation requested \\ from simulator\strut};

\draw (5.5,0) node[below, align=center]{Simulator renders\\ observation};

\draw  [thick, ->] (7,-1) -- (7, 1) node[align=center, above] {Observation sent \\ back to agent\strut};

\draw (8,0) node[above, align=center]{Agent\\ deliberates\strut};

\draw [thick, <-] (9,-1) node[below]{$A_{t+1}$} -- (9, 1) node[align=center, above] {Agent sends \\ action\strut};

\draw (10.5,0) node[below, align=center]{(Optional)\\ AndroidEnv waits};

\draw [thick, <-] (12,-1) node[below]{$O_{t+2}$} -- (12, 0) node[align=center, above] {Observation requested \\ from simulator\strut};
\end{footnotesize}

\draw [dotted] (4,-1.35) -- (4,-1.5) -- (12, -1.5) -- (12, -1.35);
\draw (9,-1.5) node[below]{$\Delta t$};

\end{tikzpicture}
    \caption{Timeline of the real-time interaction between an agent and AndroidEnv.}
    \label{fig:timeline}
\end{figure}

We note that a step time with high variance could cause unpredictable interactions with the device. For instance, an unexpectedly long agent deliberation time could turn an intended \emph{tap} gesture into a \emph{long press}. To prevent these issues, AndroidEnv can optionally insert a wait time before requesting observations, in order  to be  closer to a fixed rate of interaction, while still providing the agent with the most recent observation possible. Figure~\ref{fig:timeline} shows how the agent-environment cycle unfolds in time. Given a desired $\verb+max_steps_per_second+$, AndroidEnv waits $\Delta t = 1/ \verb+max_steps_per_second+$, in order  to come as close as possible to the desired interaction rate. The optional wait has a stabilizing effect on the time $\Delta t$ between consecutive observations when the variance in the agent deliberation and/or rendering time is large. A well-chosen step rate  can also extend the effect of a particular action, hence regularizing the corresponding training data.

\paragraph{Wrappers.} \label{wrappers} We also provide environment wrappers to help users customise their experiments. They allow modifying the observation space (e.g. \verb+ImageRescale+ wrapper to resize pixel observations), the action space (e.g. \verb+DiscreteAction+ wrapper to discretise the hybrid action space), or the interface (e.g. \verb+GymWrapper+ for agents expecting an OpenAI Gym interface).

\section{Relevant Work: Other RL Research Platforms} \label{sec:rl}

We designed AndroidEnv to complement existing platforms, by leveraging Android's rich ecosystem. In this section, we give an overview of some of these alternatives and highlight their features in relation to AndroidEnv. A summary of the features of these different platforms is given in Table~\ref{table:summary}.

\paragraph{Atari2600.~\citep{Bellemare_2013}}

This test bed was the first platform allowing an RL agent to interact with various tasks through the same observation-action interface. It allowed building agents that use a \textit{single} neural network architecture over a suite of 57 games. It has since been used as a core deep reinforcement learning research platform, leading to significant advances in algorithm design. Some of its characteristics include: a relatively small action space (18 discrete actions), operating in lock-step (i.e. the underlying simulation waits for the agent to act), and a diverse set of tasks that test core agent capabilities such as exploration, credit assignment, and generalisation. Still, the platform has limited flexibility: fetching rewards from games required environment developers to access privileged memory, the games are deterministic, the platform itself does not provide auxiliary signals to aid learning, and designing new tasks, although possible, is not easy. This is an important drawback, as the platform could be quite limiting when testing algorithms for scalability, large or continuous action spaces, stochastic or real-time environments, complex memory capabilities or language skills~\citep{machado2017revisiting}. AndroidEnv and OpenAI universe, which we discuss below, are alternatives that address of these limitations. In fact, OpenAI Universe includes Atari 2600 games, hence making the platform available for testing more complex action interfaces and real time interaction.

\paragraph{DeepMind Lab.~\citep{beattie2016dmlab}}

Deepmind Lab is a 3D environment that provides a suite of challenging navigation and puzzle-solving tasks for learning agents. 
The observation consists of a first person pixel-based view of the 3D world, along with depth and velocity information. Users can customise the resolution of the observatios, which are rendered by a GPU or by a CPU. The action interface consists of multiple simultaneous actions to control movement (translation, rotation, jump/crouch). The suite includes several task types such as resource collection, navigation and laser tagging. Although researchers can easily extend the task suite with Deepmind Lab tools for level creation, the tasks are all within this consistent 3D world. AndroidEnv tasks are not restricted to a specific world simulation, as tasks can be defined on any app or service running within the OS.

\paragraph{Minecraft.~\cite{johnson2016malmo}}

Minecraft is one of the most popular video games and an RL domain has been constructed on top of it, which raises important research challenges, due to the need for active perception and lifelong learning solutions. In this game, the agents’ success depends on their ability to navigate, build structures, interact with objects, collect resources, and avoid obstacles and other attacking entities (e.g. zombies)~\citep{minecraft14}. The platform provides a set of tools that facilitate in-game design to study and develop algorithmic solutions for specific cognitive faculties~\citep{tesser2017}. For example, recent work demonstrated that Minecraft can be a useful platform for research related to robotics, with strong support for an experimental setup based on the Object Oriented Markov Decision Process (OO-MDP) paradigm~\citep{Aluru2015MinecraftAA}.

Despite the fact that Minecraft is an open-world game with complex game-play elements that require long-term credit assignment,
RL research on Minecraft to date has been rather limited, with a strong focus on toy tasks with short horizons, restricted navigation, movement restricted to 2D, or interaction  limited to a small set of objects~\citep{oh16,tesser2017, Bonanno2016SelectingSU}. Other methods leverage prior human demonstration or prior  knowledge~\citep{guss2019minerl, Abel2015GoalBasedAP, AbelADKS16, shu2017minecraft, Frazier2019}. Moreover, the tasks are commonly designed to allow agents to act by using  images downsampled to 84 x 84 pixels as input, similar to the Atari Learning Environment. The agent is also limited to choosing from a small set of actions (e.g. 6 to 8 actions for navigation, pickup, breaking, placing, etc.) corresponding to low-level actuators that interact with the emulator.

\paragraph{Robotics/dm\textunderscore control~\citep{tassa2020dmcontrol}.}

Practitioners commonly use physical robots or run computer-based simulations for RL research on robotics. Physical devices provide the highest possible fidelity to real world problems, but they are generally costly, slow, non-deterministic and inflexible. Computer-based simulations cannot match their physical counterparts in fidelity (i.e. there is always a simulation gap), but they can scale to thousands or even millions of instances at a fraction of the cost. This is important for RL research due, because  RL algorithms can be data inefficient. Moreover, defining rewards that match the expectations of designers can be particularly challenging in robotics. The most common approach to overcome both of these challenges is to rely on human demonstrations. 

MuJoCo \citep{6386109} is a widely used simulator in RL research, and the basis of dm\textunderscore control, a suite of various robotics-like tasks. Its observations and actions are sets of continuous multidimensional vectors, and they vary according to different body types (e.g. humanoid, quadruped, half-cheetah etc). Users can conveniently pause and resume  the simulation of the environment at will. Moreover, tasks are easily modifiable by customising XML task descriptions, and they can be easily inspected by using physical interactivity tools provided by the engine.

\paragraph{OpenAI Universe.~\citep{open_ai_universe}} 

The Universe platform, released in 2016, has the same broad goals and motivation as AndroidEnv. Both platforms expose similar universal visual interfaces, i.e. pixels for observations. Universe provides keyboard and mouse gestures for actions. Moreover,  both platform allow for the easy design and addition of a wide variety of tasks, and the incorporation of auxiliary structured information.However, Universe predominantly specifies the reward function through a convolutional neural network that extracts numbers from images, while AndroidEnv has access to app logs and system events to compute rewards.

Universe was in many ways ahead of its time. State of the art RL agents at the time of its release were not even close to addressing all the challenges that the environment offered. Universe included Atari games in its task suite, yet no agent could adequately play them using the Universe interface, i.e. mouse and keyboard gestures and large observations. To demonstrate learning, the authors discretised the action interface and specialised it to select only among a fixed number of keyboard keys that would fully control the Atari Suite. As shown in the empirical results, AndroidEnv presents a variety of tasks, some which are definitely within reach for current RL agents, and some which are quite challenging, therefore providing an interesting runway for novel RL agents.  
\paragraph{World of Bits (WoB) \citep{pmlr-v70-shi17a}.} WoB is an RL environment based on OpenAI Universe, with tasks defined on real or cached web pages from the internet. The observation contains pixels and the Document Object Model (DOM) of the current page, along with useful annotations such as bounding boxes of DOM elements. Much like Universe, keyboard and mouse events determine the action space, inheriting its universal applicability. Users can handcraft WoB tasks or collect them via crowd-sourcing Question-Answer interactions. In particular, WoB and MiniWob++ \citep{liu2018reinforcement} include a variety of tasks that expose user interface challenges for RL agents based on similar interactions with a single Android application.

\begin{table}[h]
	\centering
	\caption{Summary of environment properties}
	\vspace{-10pt}
	\begin{center}
	\begin{tabular}{ |p{30mm} || P{22mm}| P{22mm}| P{22mm}| P{22mm}|}
		\hline
		Environment & Universal \newline Interface & Extensible Task Suite  & Real-time & Continuous Action Space\\
		\hline \hline
		Atari   &  \cellcolor{gray!40} \xmark   &     &    & \\
				\hline

		DM Lab  &   \cellcolor{gray!40} \xmark    &   \cellcolor{gray!40} \xmark    &    &  \cellcolor{gray!40} \xmark \\
        \hline

		DM Control Suite    &      & \cellcolor{gray!40} \xmark   &   & \cellcolor{gray!40} \xmark   \\
		\hline
		Minecraft &  & \cellcolor{gray!40} \xmark  & & \cellcolor{gray!40} \xmark \\
		\hline
		OpenAI Universe &    \cellcolor{gray!40} \xmark   &  \cellcolor{gray!40} \xmark   & \cellcolor{gray!40} \xmark  & \cellcolor{gray!40} \xmark \\
		\hline
		World of Bits &    \cellcolor{gray!40} \xmark   &  \cellcolor{gray!40} \xmark   & \cellcolor{gray!40} \xmark  & \cellcolor{gray!40} \xmark \\
		\hline
		\textbf{AndroidEnv}  &   \cellcolor{gray!40} \xmark    &   \cellcolor{gray!40} \xmark  & \cellcolor{gray!40} \xmark  & \cellcolor{gray!40} \xmark \\
		\hline
	\end{tabular}
	\vspace{-20pt}
	\end{center}\label{table:summary}
\end{table}

\section{Conclusion}

We described AndroidEnv, an AI platform based on the Android Operating System, which provides tasks based on its large app ecosystem. The environment's universal observation and action space, along with real-time simulation, make it a particularly interesting challenge for current state-of-the-art agents. AndroidEnv is a suitable environment for studying a wide range of RL research problems such as exploration, hierarchical RL, transfer learning, or continual learning. We hope that it will provide a useful complement to the existing set of research platforms.
 Since Android has billions of users, and AndroidEnv provides tasks that run on the standard Android OS simulator, agents trained on the platform could potentially tackle a wide range of use cases leading to direct, real-world impact. For example, the ability to automatically learn sequences of actions might lead to advanced hands-free voice navigation tools; on-device AI models could help provide a better user experience; and trained agents could assist in device testing and quality assurance by benchmarking new apps, measuring latency, or detecting crashes or unintended behaviours in the Android OS.

\bibliographystyle{abbrvnat}
\setlength{\bibsep}{5pt} 
\setlength{\bibhang}{0pt}
\bibliography{main}

\section*{Acknowledgements}
It would have been impossible to create AndroidEnv without the help of many people:

\begin{itemize}
    \item Mathieu Méa for integrating dozens of open source apps.
    \item Natalie Lambert for coordinating between research teams and interfacing with external parties.
    \item Xutong Zhao for setting up benchmarks and sandboxing.
    \item Daniel Rodriguez, Gabriel Taubman, Chris Rawles, Wei Li, Robert Berry and Alice Li for extensive research collaborations with Google.
    \item Eser Aygün, Rui Zhu, Tom Ward, Alexandre Moufarek, Jay Lemmon, Davide Vercelli,
    Alban Rrustemi, Jeff Stanway, Damion Yates, David Barker, Duncan Williams, Tim Harley and Erwin Jansen for helping to set up AndroidEnv on Google's infrastructure.
    \item Linfeng (Frank) Yang and Yahan Zhou for Android Emulator guidance.
    \item Justin Novosad,  Antonio Maiorano, Sean Risser, André Kaba, Alban Chagnoleau, Loïc Gelle, Jing Liu and Félix Larose-Gervais for prototyping interesting ideas in the early stages of the project.
    \item Tom Ward, Ankit Anand and Nando de Freitas for very useful feedback on a draft of this report.
    \item Phoebe Kirk, Gabrielle Ohlsen, Michelle Dunlop, Richard Ives for their legal advice.
    \item Aliya Ahmad, Emma Yousif, Louise Deason, Malcolm Reynolds for assistance on the open sourcing process and on communications.
    \item The Google and DeepMind Montr\'eal team for enthusiastic discussions throughout the inception and refinement of AndroidEnv.
\end{itemize}

\end{document}